# Smoothing the Rough Edges: Evaluating Automatically Generated Multi-Lattice Transitions

Martha Baldwin,[1] Nicholas A. Meisel,[2] Christopher McComb[1]

[1]Department of Mechanical Engineering, Carnegie Mellon University, Pittsburgh, PA 15213
[2]School of Engineering Design and Innovation, The Pennsylvania State University, University Park, PA 16802

## Abstract

Additive manufacturing is advantageous for producing lightweight components while addressing complex design requirements. This capability has been bolstered by the introduction of unit lattice cells and the gradation of those cells. In cases where loading varies throughout a part, it may be beneficial to use multiple, distinct lattice cell types, resulting in multi-lattice structures. In such structures, abrupt transitions between unit cell topologies may cause stress concentrations, making the boundary between unit cell types a primary failure point. Thus, these regions require careful design in order to ensure the overall functionality of the part. Although computational design approaches have been proposed, smooth transition regions are still difficult to achieve, especially between lattices of drastically different topologies. This work demonstrates and assesses a method for using variational autoencoders to automate the creation of transitional lattice cells, examining the factors that contribute to smooth transitions. Through computational experimentation, it was found that the smoothness of transition regions was strongly predicted by how closely the endpoints were in the latent space, whereas the number of transition intervals was not a sole predictor.

# 1. Introduction

Additive manufacturing (AM) is known for its ability to produce lightweight components while addressing complex design requirements. This ability has been bolstered by the introduction of unit lattice cells, which enable designers to significantly reduce the weight of a part while maintaining necessary stiffness and required geometry.[1,2] Proposed applications of lattices are often centered around lightweighting[3,4] or improving the stiffness-to-weight ratios of a component.[1] Lattices are also useful in impact reduction applications,[5] such as the crumple zones in vehicles[1] or surgical implants that experience variable loading.[6,7] Graded lattices[8–12] vary the thickness of individual strut-based unit cells to accommodate areas of high or low stress, introducing further design freedom and outperforming the stiffness of uniform lattice structures of comparable weight.[9,10,12] Given the progressive deformation properties of graded lattices[11], it is believed that they would be more beneficial than uniform lattices in impact reduction applications.[13] The introduction of multi-lattice structures, which are structures comprised of multiple types of unit cell topologies, introduces even more design freedom.[6,14–17] Similar to graded lattice structures, multi-lattice structures have higher strength and stiffness than uniform lattice structures of comparable density.[13,16] However, the transition regions between lattice types in multi-lattice structures must be carefully designed to avoid stress concentrations between different unit cell topologies, since such concentrations are detrimental to overall part strength.[16] Unfortunately, designing such complicated mechanical transition regions is challenging for human designers.

To address this challenge, recent work has utilized machine learning to rapidly design transition regions between unit cells.[18–20] However, there has been little quantitative assessment of the geometric smoothness of the resulting transitions. In this work, we create a Variational Autoencoder (VAE) system that can design transitions between different unit lattice cells by interpolating in the latent space. Through interpolation, transition regions can be produced with any number of intervals between two end topologies, due to the generative nature of a VAE latent space. Much like prior work, this system can provide users with a new design opportunity that could serve to expand the design methods available in AM.[21] In order to determine the effectiveness of VAEs for producing transition regions, and in contrast to prior work, our primary research questions address the unknowns about the effects of dimensionality reduction with respect to lattice topologies:

1. How does distance in the latent space effect the smoothness of interpolations?
2. How does the number of transition intervals affect the smoothness of interpolations?

In response to the first research question, we hypothesize that the smoothness of interpolations will decrease as distance in the latent space increases. This hypothesis is based on the premise that the images should be less similar to one another the farther apart they exist in the latent space, leading to a greater disparity between images and a more challenging interpolation. Our second hypothesis is that the smoothness of interpolations will increase as the number of transition intervals increases. This is based on the intuition that increasing the number of transition intervals decreases the distance between each step in the transition.

## 2. Background
### 2.1 Unit Cells for Design

This work examines periodic lattice structures, which are created by repeating a unit cell. A *uniform lattice* structure consists of one type of unit cell, where all the cells in the structure have the same density and size.[12] The properties of each type of unit lattice are unique,[1,4,5] making it important to select the correct type of lattice for a design. A *graded lattice* structure consists of one type of unit cell, but the cells in the structure can vary in volume fraction.[10] It has been proven that functionally graded lattice structures can achieve higher stiffness than uniform lattices[9,10,12], taking advantage of a higher degree of design freedom.[11,12] Researchers have suggested that such properties would be beneficial in applications where there are dynamic loads, such as surgical implants[6] or impact protection equipment.[13] Although these structures seem relatively simple to create, they can still be challenging to comprehend and design. The difficulty of designing these structures can be mostly attributed to the complexity of the variable lattice shapes, and the lack of robust DfAM software. Li et al. used relative density mapping from topology optimization and assigned densities based on stress requirements.[8] However, the key restriction in graded lattice structures is their property dependence on the type of unit cell. It was found that surface-based unit cells outperformed strut-based cells due to their connectivity,[12] as well as the degree of gradation greatly affecting the cumulative energy absorption for body-centered cubic lattices, but not Schwarz-P lattices.[2] These restrictions have encouraged researchers to explore other alternatives to graded lattice topologies, such as multi-lattice topologies.

A *multi-lattice* structure is being referred to as a structure that contains multiple types of unit cells. Many studies that explore multi-lattice structures have utilized unit cells with matching boundaries to avoid developing a method to connect nodes.[6,13,16,17] One method for creating the structures is using relative density mapping from topology optimization to assign 2.5-dimensional unit cells based on stresses in the system, where the cells assigned were of various topologies.[13,16,17] As a result, these papers were able to prove that structures utilizing 2.5-dimensional multi-lattice structures often outperform uniform lattice structures regarding stiffness and strength.[13,16,17] Another study conducted by Gok showed that using a multi-lattice hip implant reduced the maximum stress and weight compared to conventional hip implants.[6] Although the work does not provide comparison between other types of mesostructured patterning, it does demonstrate a strong use case for multi-lattice structures. It should be noted that the study was also restricted to unit cells that had overlapping boundaries, so the transition regions were inherently smooth.

Although multi-lattice structures have demonstrated utility in certain situations, there are also potential drawbacks. Kang et al. concluded that the boundaries between unit cell types caused stress concentrations if not carefully designed, decreasing overall part strength.[16] This indicates that there is a need for smooth transitions between unit cells to appropriately distribute stress through structures. A numerical approach to solving this problem was executed by Sanders et al., who created transition regions by using signed distance functions with respect to the boundaries of each strut.[14] Although this technique can be applied to "unit cells composed of noncylindrical bars or plates", the interpolations between each pair of unit cells must be computed individually. Although this method can produce smooth transition regions in multi-lattice structures, for those looking to optimize both the macrostructure and the mesostructure, this method is extremely repetitive and tedious. Wang et al. have also regarded the creation of transition regions as "a challenging problem involving complex inverse design at the microscale, costly nested optimization at the macroscale, and boundary matching between neighboring microstructures."[18] These works underscore the criticality of being able to create smooth transitions between unit cells using machine learning methods.[14]

## 2.2 Machine Learning Methods

Recent approaches to creating continuous transitions in multi-lattice structures have utilized machine learning. Although a wide variety of potential approaches exist, this work focuses on two methods that have been used predominantly in the literature for creating multi-lattice structures. Specifically, this section explores: Generative Adversarial Networks (GANs) and Variational Autoencoders (VAEs).

**Generative Adversarial Networks**

Generative Adversarial Networks (GANs) utilize two models that work in parallel to ultimately create a single model for generating data.[22] These two models are known as the generator and discriminator, and they are adversarial in nature. Specifically, the generator learns to produce fake data based on a training set of real data, while the discriminator learns to distinguish between the fake and real data. As training progresses, each model progressively becomes better at its purpose until the discriminator is unable to distinguish between real and fake data produced by the generator. This can be used to generate a wide variety of outputs, including images[22], voxels for topology optimization[19], and meshes for computational fluid dynamic simulations.[23]

Wang et al. used an Inverse Homogenization GAN to generate a multi-lattice structure that reduced stress concentrations by nearly 80%.[19] This serves as a contributing motivation for the current work, as it proves there is a benefit to creating multi-lattice structures. However, the primary drawback to this method was that it was only intended to organize the types of unit cells, but only based on performance rather than shape. Therefore, interfaces between the cells were not perfect, and a second interpolation had to be performed to create smooth transition regions. Again, this highlights the difficulties involved in designing effective transition regions.

**Variational Autoencoders**

A variational autoencoder (VAE) is a machine learning model that learns how to perform data-driven dimensionality reduction.[24] Reducing the dimensionality of data can make it more computationally efficient to perform inferences, such as interpolations. What makes VAEs unique from other techniques is that they perform a non-linear dimensionality reduction, which is advantageous for reducing complex data types.[25] The reduced data is combined to establish the reduced dimensionality latent space, which can be used to represent complex mechanical parts[26] and performance characteristics of engineered systems.[27] Due to the dimensionality reduction techniques of VAEs, the latent spaces they produce will be non-Euclidean.[28]

A VAE consists of two models, an encoder and a decoder, which work in series. The encoder performs a dimensionality reduction operation to create a latent representation of the data.[22,25] The decoder uses the information from the latent space to reconstruct the data. Finally, the VAE is trained by minimizing the error between the reconstructed output from the decoder model and the original data provided. VAEs are appealing for the current application since they have appeared in similar work. For instance, Wang et al. constructed a VAE to perform multi-lattice interpolations between 2D lattices.[18] We extend that work by critically interrogating the relationships that exist within the VAE latent space.

## 3. Methodology

In order to simplify the data needs and reduce the training time of the VAE model, this work primarily addresses the design of 2-dimensional lattices. The methodological approach is shown in Figure 1, with each block aligning with part of this section. Subsection 3.1 will describe how artificial data was generated to meet testing needs. Subsection 3.2 outlines the hyperparameters and key characteristics of the VAE used for training. Subsection 3.3 shows the process for creating interpolations in the latent space. Subsection 3.4 will outline the experimental process to measure the performance of transition regions.

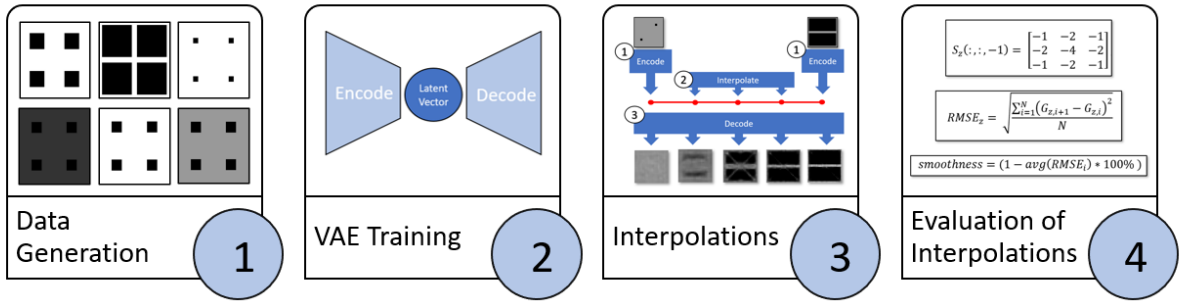

Figure 1: Illustration of overall methodologies

### 3.1 Synthetic Data Generation

Synthetic data was generated based on a series of 12 shapes (see Figure 2) that mimicked strut-based lattices found in the additive manufacturing literature.[13,17] To add variation to the data, the thickness of the struts and the density of pixels were varied (see Figure 3). The density represents the value of each pixel between 0 (black, representing material absence) and 1 (white, representing material presence). This representation of density aligns with that used in the topology optimization literature and is useful here for developing a denser latent space. Once all the shapes were generated, there were a total of 415 data points.

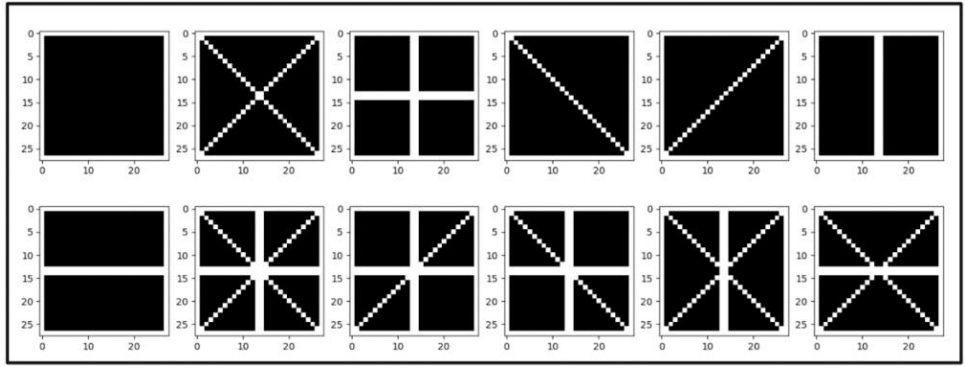

Figure 2: Synthetic Shape Types

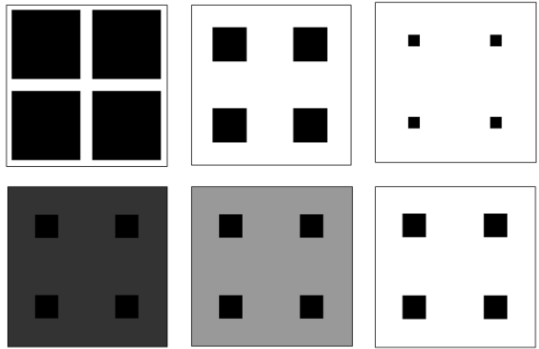

Figure 3: Original shape with possible density and pixel intensity variations

### 3.2 Training the Variational Autoencoder

The architecture of the VAE used in this work is designed to mimic an architecture used in prior work (see Figure 4).[29] Perhaps the most important hyperparameter of a VAE is the dimensionality of the latent space. A qualitative study of latent space dimensionality revealed that a dimensionality of 4 provided a desirable balance between training time and reconstruction accuracy.

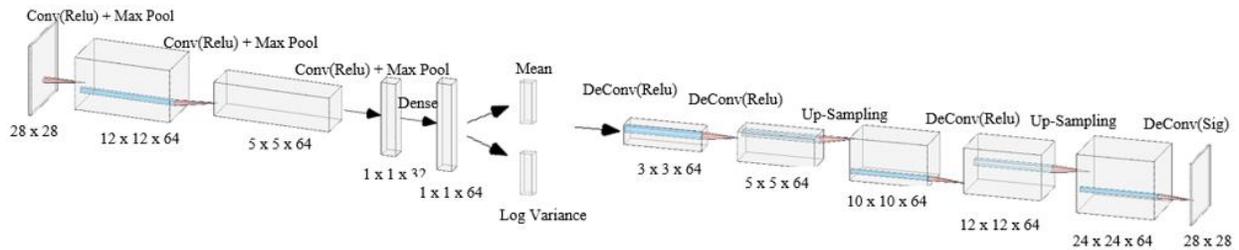

Figure 4: Framework of VAE[*]

The VAE was trained using a batch size of 32 and the Adam optimizer.[30] 85% of the dataset was used to train the VAE and 15% was reserved to validate it. Training was terminated early if the loss failed to improve after 10 epochs and the variational autoencoder was also designed to save the weights from the iteration with the best validation loss.

### 3.3 Creating Interpolations

Once trained, the VAE has established a latent space in which interpolations can be performed (See Figure 5(a)). The encoder can be used to map lattices into the latent space, while the decoder can be used to map from the latent space back to a lattice. The unit cells at the end of the desired transition region are first provided to the encoder, which calculates the latent points of the two cells (see step 1 in Figure 5(b)). The transition intervals are then interpolated between the encoded endpoints in the latent space (See step 2 in Figure 5(b)). Finally, the decoder generates the cells that correspond to those latent points (See step 3 in Figure 5(b)). It should be noted that the decoder is generating new lattice cells based on the latent points provided. For this work, we chose to use linear interpolations to generate a simple proof of concept. As such, it should be noted that the distance in some latent spaces should not be measured linearly.[28,31] However, the results from this work indicate that this method of interpolation was sufficient, see notes in the discussion section.

---

[*] Image created with http://alexlenail.me/NN-SVG/AlexNet.html

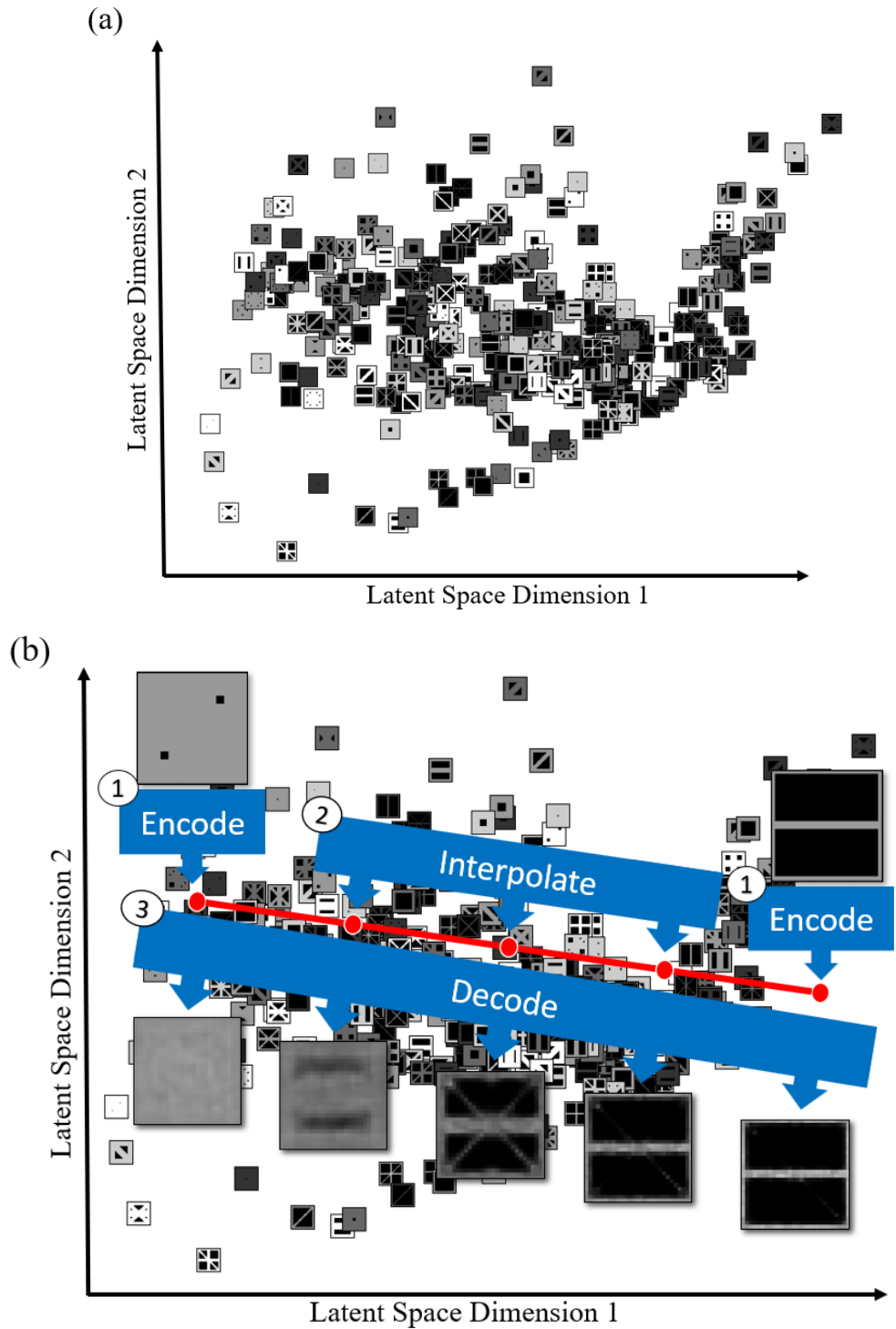

Figure 5: (a) Depiction of the latent space (b) Diagram demonstrating interpolation process in the latent space (1) Encoding of test images (2) Interpolation between the encoded endpoints (3) Decoding of all points to produce transition region. The 2D visual of the latent space was created

by performing a PCA reduction on the data, which originally existed in a four-dimensional latent space.

Figure 6 shows a visualization of the latent space and a sample interpolation, where each thumbnail represents a unit cell. This visual depicts the results from the steps outlined in Figure 5(b). Figure 6(a) demonstrates the feasability of using VAEs for creating transition regions, as it demonstrates a clear example of how a transition region could be developed over a greater number of transition intervals. By defining the scope of the transition region, the potential for a 2D multi-lattice structure is established.

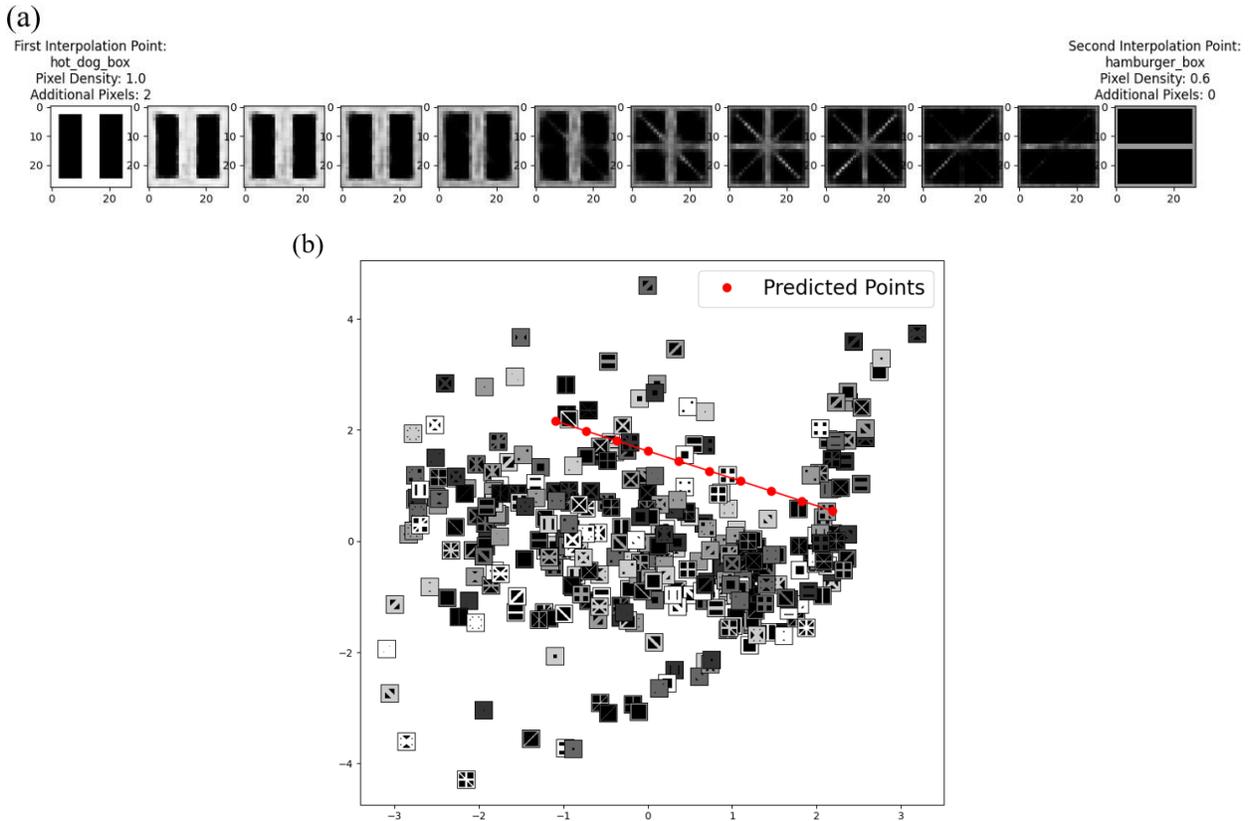

Figure 6: Demonstration of Interpolation: (a) Interpolation (b) PCA reduced latent space plot with the interpolation points superimposed in the space. The 2D visual of the latent space was created by performing a PCA reduction on the data, which originally existed in a four-dimensional latent space.

## 3.4 Experimental Evaluation

To answer the research questions introduced in this work, we design an experiment to vary parameters of the transition generation task and measure the effect on the smoothness of transition regions. In this experiment, we measure the smoothness of transition regions as we vary the distance in the latent space from -3 to 3 standard deviations from the mean, while simultaneously changing the number of transition intervals from 5, 10, to 15 points. By measuring distance in terms of standard deviations, we could ensure the entire span of the latent space was being tested with 6 different distances between transitions. The number of transition intervals were selected to give a range of different transition regions to analyze.

To accurately gauge the quality of an interpolation, a metric was designed to measure the smoothness of the transition region developed. Here, *smoothness* should (1) measure how the transition region changes both within images and across images, (2) penalize pixels disconnected from the main structure, as we did not want to encourage the generation of floating pixels, and (3) account for pixel intensity, as interpolations will often produce non-binary images. Related metrics have been published in literature[19,20,31–34] but each of these metrics fails to satisfy at least one of these criteria. We create a new metric that emphasizes change along the edges of the unit cells using a 3-dimensional Sobel filter.[35,36] This makes it possible to resolve both within image changes (here the $x$ and $y$ directions) and between-image changes (here the $z$ direction) using a single operation.[34] Specifically we make use of Amin et al.'s formulation of 3-dimensional Sobel filters, $S_x$, $S_y$, and $S_z$, which are the Sobel filter components in the x, y, and z directions respectively.[36] From an additive perspective, the goal of this metric is to encourage smooth geometry for manufacturability.

$$\begin{aligned}
G_{x,i} &= S_x(:,:,-1)conv(I_i) + S_x(:,:,0)conv(I_{i+1}) + S_x(:,:,1)conv(I_{i+2}) \\
G_{y,i} &= S_y(:,:,-1)conv(I_i) + S_y(:,:,0)conv(I_{i+1}) + S_y(:,:,1)conv(I_{i+2}) \\
G_{z,i} &= S_z(:,:,-1)conv(I_i) + S_z(:,:,0)conv(I_{i+1}) + S_z(:,:,1)conv(I_{i+2})
\end{aligned} \quad (1)$$

where $G_x, G_y,$ and $G_z$ are the gradient array components in the $x$, $y$, and $z$ directions respectively, $I$ is the image, and $i$ is the index between each gradient array and their respective images.

To directly compare the gradients, the $x$, $y$, and $z$ components of the arrays were flattened. Then the root mean squared error (RMSE) between the consecutive gradients was measured, as that would be the indicator of "smoothness" between each dimension (Eq 3).

$$\begin{aligned}
RMSE_{x,i} &= \sqrt{\frac{\sum_{j=1}^{N}(G_{x,i+1,j} - G_{x,i,j})^2}{N}} \\
RMSE_{y,i} &= \sqrt{\frac{\sum_{j=1}^{N}(G_{y,i+1,j} - G_{y,i,j})^2}{N}} \\
RMSE_{z,i} &= \sqrt{\frac{\sum_{j=1}^{N}(G_{z,i+1,j} - G_{z,i,j})^2}{N}}
\end{aligned} \quad (2)$$

where $RMSE_x, RMSE_y, and\ RMSE_z$ are the root mean squared errors of a pair of gradients in the $x$, $y$, and $z$ directions respectively, $j$ is the index that identifies the specific term in the gradient array, and N is the number of terms in a single gradient array.

Once the RMSE was calculated for the $x$, $y$, and $z$ components, the average RMSE was calculated and normalized to create a final "smoothness value".

$$RMSE_i = \frac{RMSE_{x,i} + RMSE_{y,i} + RMSE_{z,i}}{3 \cdot RMSE_{max}} \qquad (3)$$

where $RMSE_{max}$ is the maximum possible $RMSE_i$ which is calculated based on the filter used. The normalization of the RMSE allowed for the smoothness value to be represented as a percentage.

$$smoothness = (1 - avg(RMSE_i)) \cdot 100 \qquad (4)$$

More details on the implementation of this smoothness evaluation are available in prior work by the authors.[37]

## 4. Results

Overall, the purpose of this experiment is to investigate the properties related to smoothness of VAE-generated multi-lattice transition regions. To test the research questions and hypotheses introduced previously, a variety of different interpolations were defined and assessed. Following the experimental design outlined previously, we produced transition regions to test our hypotheses. A sample of interpolations with increasing distance in the latent space can be seen in Table 1.

Table 1: Visualization of transition regions based on number of standard deviations with 10 transition intervals

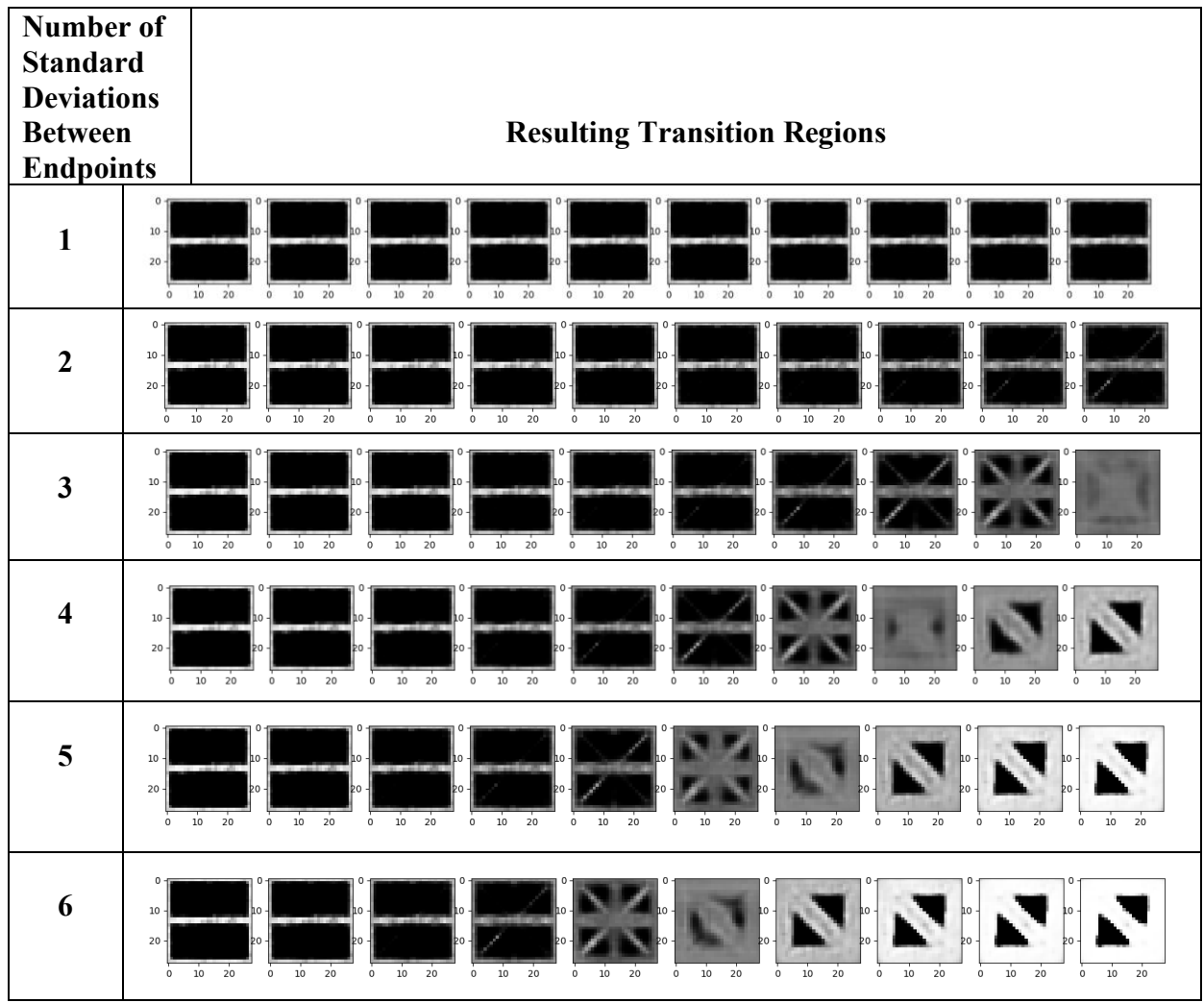

Visually, the unit cells are most similar when there is 1 standard deviation between the endpoints. To confirm this intuition, we conducted the complete experiment described in the previous section (see Figure 7).

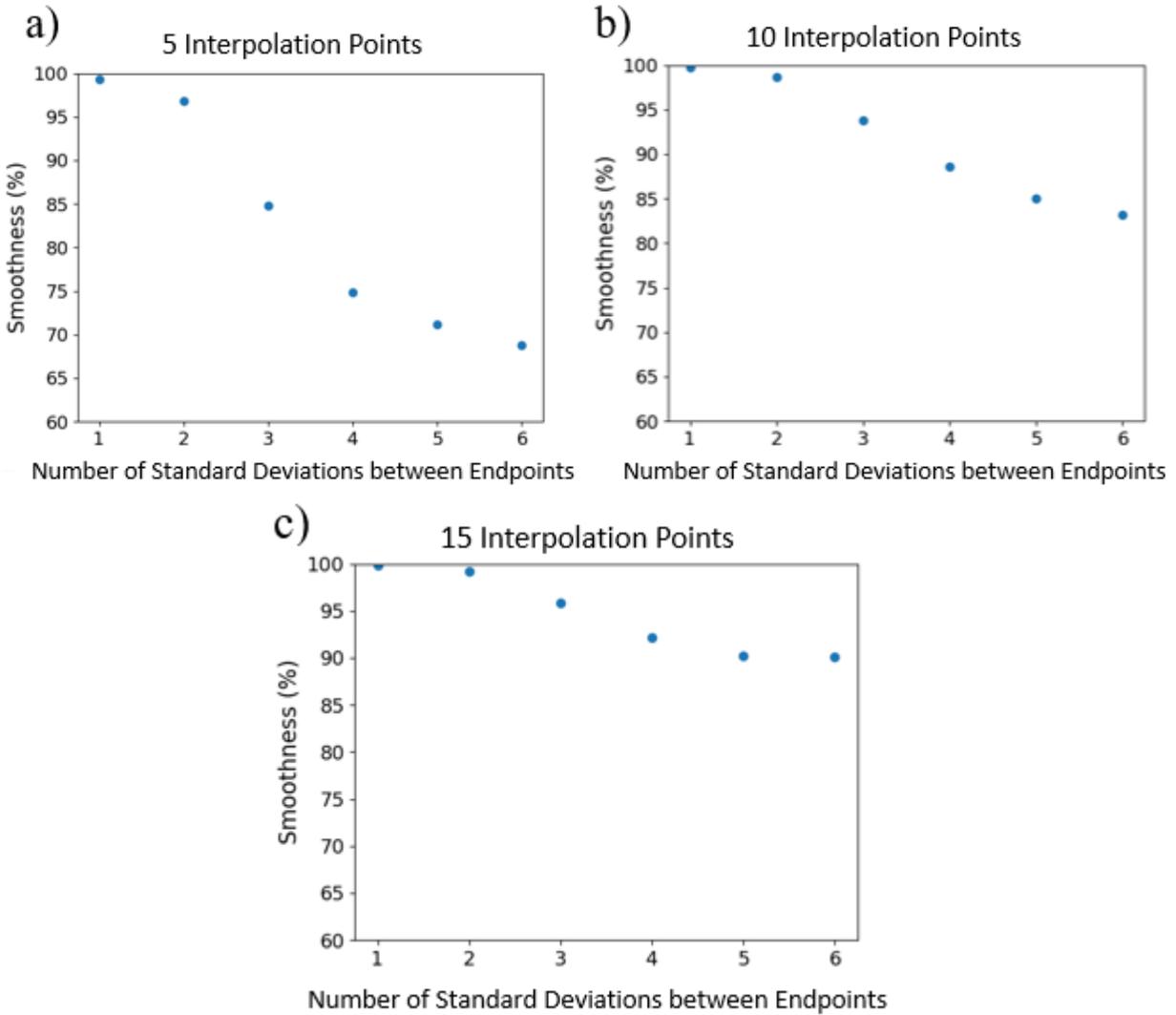

Figure 7: Smoothness vs Number of Standard Deviations in the Latent Space. (a) Analysis based on 5 interpolation points - Fewer interpolation points result in significantly rougher transitions over a longer distance in the latent space. (b) Analysis based on 10 interpolation points (c) Analysis based on 15 interpolation points - More interpolation points in the latent space result in slightly poorer transitions as distance increases.

To confirm the results, an ordinary least squares (OLS) regression was executed on the data gathered in Figure 7, where the dependent variable was smoothness, and the independent variables were the number of transition intervals and distance in the latent space. This regression was found to be highly statistically significant ($p < 0.001$) as well as practically significant ($R^2 = 0.930$) (see Table 2).

Table 2: Ordinary Least Square Regression

|  | Coefficient | Standard Error | p |
|---|---|---|---|
| Constant: | 105.8102 | 3.777 | p < 0.0001 |
| Number of Standard Deviations (Distance): | -7.8790 | 0.970 | p < 0.0001 |
| Number of Transition Intervals: | -0.2605 | 0.350 | 0.469 |
| Interaction Term: | 0.4087 | 0.090 | p < 0.0001 |

It was hypothesized that greater distance in the latent space would result in a decrease in smoothness as points become spread further apart. This hypothesis was based on the premise that the images should be less similar to one another the farther apart they exist in the latent space, since it is organized based on similarity. The p-values in Table 3 confirmed that as the distance between endpoints increases, the smoothness decreases. The magnitude of the p-values also indicates that distance in the latent space has the most significant effect on smoothness.

It was further hypothesized that the number of transition intervals would have a proportional relationship with the smoothness of the transition regions. This was based on the idea that if there are more transition intervals, then the distance between each step in the transition region would be smaller, therefore, the unit cells should be more similar. However, the results indicate that the number of transition intervals alone is unlikely to influence the smoothness value, based on the p-value of 0.469 (see Table 2). Consequently, the combination of the two variables, noted by the interaction term in Table 2, does directly affect the smoothness. This indicates that the number of transition intervals only affects the smoothness when distance is also accounted for. Given these results, the best transition regions can be produced by reducing the distance traveled in the latent space as much as possible.

## 5. Discussion

This work underscores the potential of VAEs to support transition region design. However, the smoothness of the transition region is primarily limited by the distance in the latent space, given that the transition is linear. Therefore, our hypothesis must be partially refuted, as the number of transition intervals did not have the expected impact on smoothness. As stated previously, the data set consisted of 415 data points. Although limited, this dataset proved sufficient for the purposes of this paper. The work focused on a specific class of strut-based unit cells and had a small sample size to avoid overfitting.[20] The goal was centered around learning about how latent spaces work using an applicable metric. This was achieved, given that the interpolations in the latent space generated smooth transition regions. As mentioned in the background, the manifold of the latent spaces from VAEs are typically non-Euclidean.[28] However, based on the results in Figure 7, the space appears to have some Euclidean characteristics as the smoothness temporarily decreases linearly. This would be the expected result if the manifold was linear around the mean of the latent space.

This section further explores the effects of the qualitative differences in unit cell shape with respect to smoothness. It is believed that as the endpoints of an interpolation become more "different", the smoothness of the transition region should decrease. Given the complexity of the latent space, visual inspection may be deceptive, which is why it is important to address the unexpected. Table 3 consists of an array of various transition regions, all of which have the same initial interpolation point. Although the distances between the endpoints are not the same, this series of transition regions are ordered based on smoothness.

Table 3: Examples of different topologies and their effect on smoothness of transition regions

| Example | Example Transition Regions | Smoothness (%) |
|---|---|---|
| 1 | 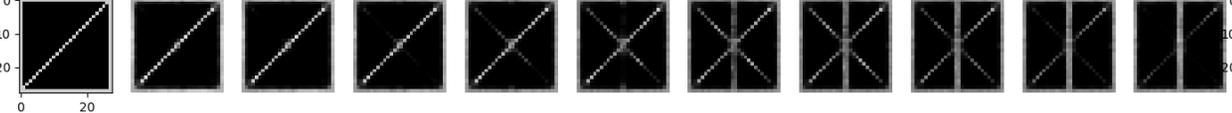 | 95.946 |
| 2 | 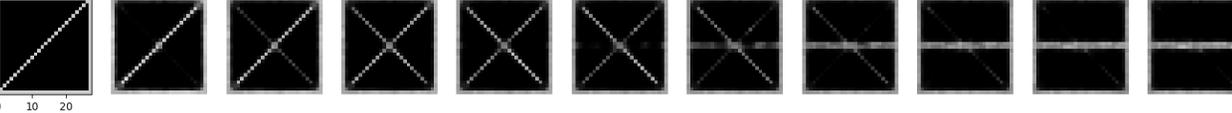 | 95.398 |
| 3 | 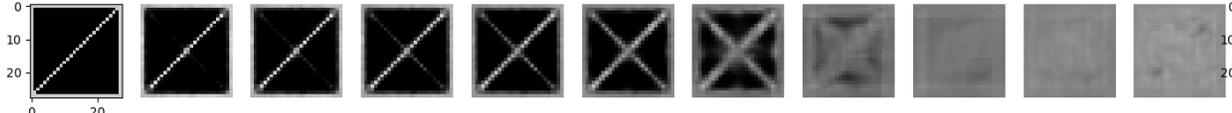 | 91.576 |
| 4 | 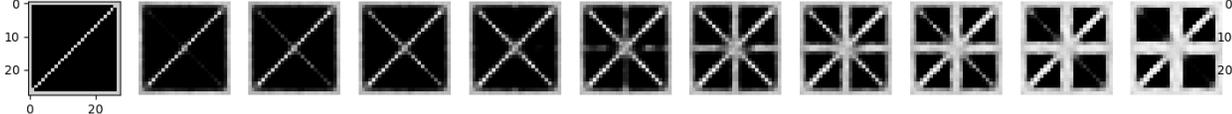 | 90.928 |
| 5 | 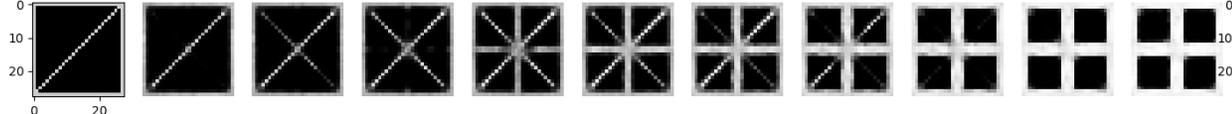 | 88.141 |

From a qualitative standpoint, the rankings were expected to order Example 1 > 2 > 4 > 5 > 3, based on the similarity of the endpoints. However, the smoothness metric is based on the variation along all the unit cells, not solely the endpoints. In other words, it is a path-based measure, relying on all intermediate frames; whereas human intuition is more point-based, relying instead predominantly on the endpoints. Example interpolations 1 and 2 are similarly smooth, as both require a "rotation" of an element within the cell. Example 3 was less smooth than Examples 1 and 2, but there was a large number of pixels that needed to be activated. However, based on the qualitative evaluation, Example 3 should have performed the worst. Since the smoothness evaluation is highly dependent on the amount of similarity between consecutive unit cells, the endpoint unit cells in Example 3 would have increased the smoothness percentage significantly. Finally, Example 4 performed slightly better than Example 5, as the diagonal was not removed in Example 4, meaning that fewer changes occurred.

It should be noted that the methodology explored in this work is not limited to interpolations along lines in the latent space. Figure 8(a) shows how powerful VAEs are by demonstrating a smooth transition region between four distinct unit cells, effecting a gridded interpolation. Such a structure represents how a transition region would be developed between four different lattice topologies. From a design perspective, this ability would allow users to select unit cells with various desired physical properties to fill a structure, since the transitions between all cells are defined. Multi-lattice designs incorporating more than two types of lattices significantly expand design possibilities for creating transition regions. What makes this approach unique is its ability to accommodate the needs for a variety of users. If a user would like to use a simple graded lattice structure, then they would only need to select endpoints of the same unit cell class with different densities. In its current state, the model could be directly used for creating these structures, as there is no concern for discontinuity among those structures.

Figure 8(b) represents the locations of each of the predicted points in the PCA reduced latent space. As established previously, the closer points will have smoother transitions based on the smoothness metric. Figures 8(c) and 8(d) prove that the distance in the PCA space directly correlates to these smoothness values, meaning that as the endpoints move closer to one another in the latent space, their transitions become smoother. The columns exhibit much smoother transition regions, whereas the rows exhibit much rougher transition regions, which is expected based on their distances between endpoints. Additionally, based on a qualitative analysis of the mesh, it is clear that the columns contain more similarity than the rows. This is shown quantitatively in Figure 8(e), where the plot shows the smoothness of each row and column in the mesh.

(a)

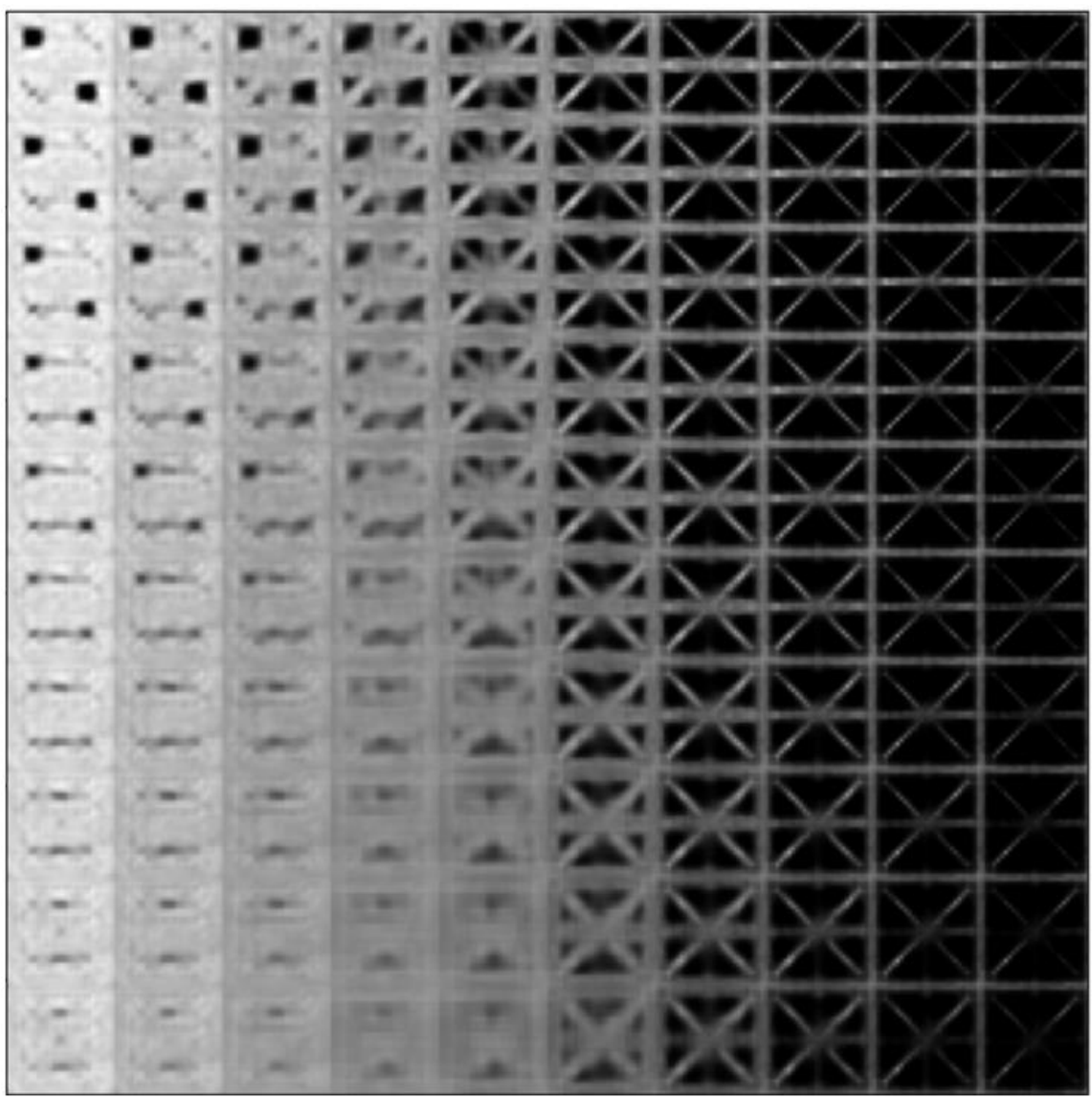

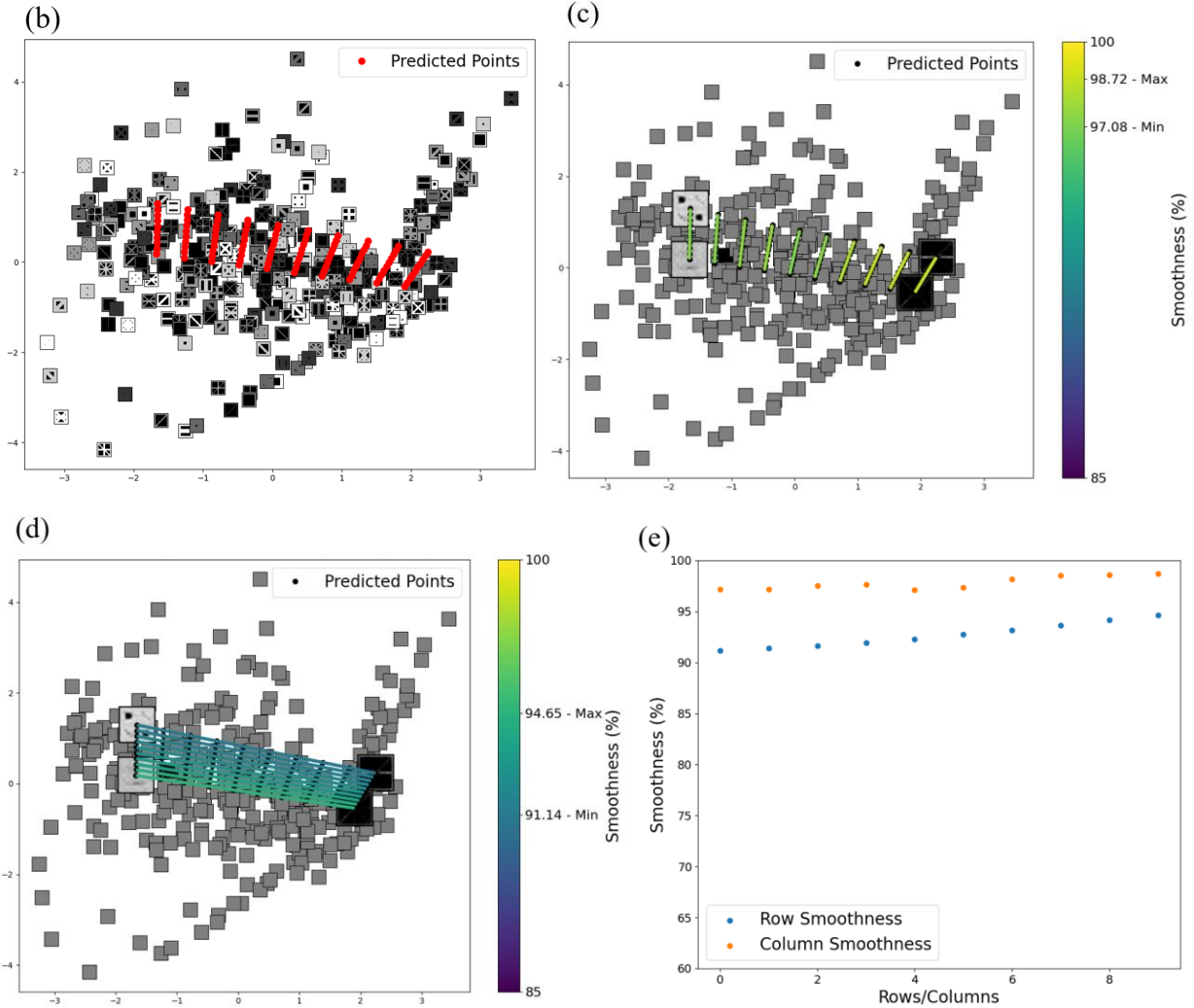

Figure 8: (a) Example of mesh grid interpolation (b) PCA reduced representation of the mesh grid interpolation in the latent space (c) Mesh grid interpolation with labeled smoothness for each column (d) Mesh grid interpolation with labeled smoothness for each row (e) Plot of Smoothness vs Rows and Columns. The 2D visual of the latent space was created by performing a PCA reduction on the data, which originally existed in a four-dimensional latent space.

## 6. Conclusion

In cases where the loading varies throughout a part, it may be advantageous to design a multi-lattice structure. However, abrupt transitions between lattice types may cause stress concentrations, making the boundary a primary failure point; thus, transition regions between lattice cell types must be carefully designed. This work demonstrates and assesses a method for using variational autoencoders to automate the creation of transitions amongst multi-lattice structures. In comparison to other computational approaches, this work focused on assessing the impact of latent space characteristics on transition smoothness.

Specifically, we examined how distance in the latent space and the number of transition intervals affects the smoothness of a transition region. We created a variety of interpolated transition regions using a VAE and then evaluated those transition regions using 3D Sobel filters. Through OLS regression, it was found that the distance between endpoints in the latent space had the most significant impact on smoothness, whereas the number of steps in the transition was unlikely to affect the smoothness alone. Given these results, the best transition regions can be produced by reducing the distance traveled in the latent space as much as possible and including a higher number of transition intervals. These conclusions were consistent as the VAE architecture was optimized, and different sets of test and training data were used. However, these conclusions are based solely on the architecture outlined above and should therefore be reevaluated when exploring future work.

The current work focused on geometric smoothness, but future work must address smoothness of physical properties and manufacturability as well. This work is limited regarding the uncertainty of mechanical behaviors for transitions produced through geometry. It is unclear whether the geometric smoothness studied here will correlate to smoothness in physical properties within a structure as well.[18] Regarding manufacturability, future work will need to account for printer restrictions that are inherent to specific types of AM processes. The major restrictions we have identified are self-supporting angles and minimum feature size limitations, which warrant future exploration to ensure printability. Another limitation of this work is the focus on 2-dimensional unit cells, which are rarely useful in practice.[38] Although limited, this work provides a roadmap for future 3-dimensional implementations. Future work should strive to incorporate an exploration of non-linear interpolations to further optimize the smoothness of the transition regions.[31] As a limitation of this work is the measure of distance using the coordinates of the latent space itself. The distance metric in future work should incorporate a measurement to address uniquely shaped latent space configurations.[28,31] Finally, there are many avenues for expanding this work to address adjacent fields related to automatically generated lattice structures. For example, applying this type of methodology to conformal lattices, which are lattice structures that conform to the boundaries of the structure, could prove beneficial.[38]

**Authorship Contribution Statement:**
Martha Baldwin: Data Curation (lead), Formal Analysis (equal), Investigation (lead), Methodology (lead), Project Administration (equal), Software (lead), Validation (lead), Visualization (lead), Writing – original draft (lead)

Dr. Nicholas Meisel: Conceptualization (equal), Supervision (supporting), Writing – review & editing (supporting)

Dr. Chris McComb: Conceptualization (equal), Formal Analysis (equal), Methodology (supporting), Project Administration (equal), Resources (lead), Supervision (lead), Writing – review & editing (lead)

**Author Disclosure Statement:**
No competing financial interests exist.

**Acknowledgements:**
This paper is based on work that was recently accepted for publication at the Solid Freeform Fabrication Symposium.[37] This material is based upon work supported by the National Science

Foundation through Grant No. CMMI-1825535. Any opinions, findings, conclusions, or recommendations expressed in this paper are those of the authors and do not necessarily reflect the views of the sponsors.